# FReSCO: Flow Reconstruction and Segmentation for low latency Cardiac Output monitoring using deep artifact suppression and segmentation.

Olivier Jaubert[1,2], Javier Montalt-Tordera[1], James Brown[3], Daniel Knight[3], Simon Arridge[2], Jennifer Steeden[1], Vivek Muthurangu[1,3].

[1.] UCL Centre for Translational Cardiovascular Imaging, University College London, London. WC1N 1EH. United Kingdom

[2.] Department of Computer Science, University College London, London. WC1E 6BT. United Kingdom

[3.] Department of Cardiology, Royal Free London NHS Foundation Trust, London. NW3 2QG. United Kingdom.

Corresponding author:

|  |  |
|---|---|
| Name | Olivier Jaubert |
| Department | Centre for Translational Cardiovascular Imaging |
| Institute | University College London |
| Address | 30 Guilford St, London WC1N 1EH |
| E-mail | o.jaubert@ucl.ac.uk |

# Abstract

**Purpose:** Real-time monitoring of cardiac output (CO) requires low latency reconstruction and segmentation of real-time phase contrast MR (PCMR), which has previously been difficult to perform. Here we propose a deep learning framework for "Flow Reconstruction and Segmentation for low latency Cardiac Output monitoring" (FReSCO).

**Methods:** Deep artifact suppression and segmentation U-Nets were independently trained. Breath hold spiral PCMR data (n=516) was synthetically undersampled using a variable density spiral sampling pattern and gridded to create aliased data for training of the artifact suppression U-net. A subset of the data (n=96) was segmented and used to train the segmentation U-net.

Real-time spiral PCMR was prospectively acquired and then reconstructed and segmented using the trained models (FReSCO) at low latency at the scanner in 10 healthy subjects during rest, exercise and recovery periods. CO obtained via FReSCO was compared to a reference rest CO and rest and exercise Compressed Sensing (CS) CO.

**Results:** FReSCO was demonstrated prospectively at the scanner. Beat-to-beat heartrate, stroke volume and CO could be visualized with a mean latency of 622ms. No significant differences were noted when compared to reference at rest (Bias = -0.21±0.50 L/min, p=0.246) or CS at peak exercise (Bias=0.12±0.48 L/min, p=0.458).

**Conclusion:** FReSCO was successfully demonstrated for real-time monitoring of CO during exercise and could provide a convenient tool for assessment of the hemodynamic response to a range of stressors.

# Introduction

Continuous assessment of cardiac output (CO) has several applications such as evaluating the response to exercise or pharmacological innervations [1–4]. The non-invasive reference standard method of measuring cardiac output is phase contrast magnetic resonance (PCMR) [5–7]. However, conventional PCMR utilizes segmented k-space acquisitions and therefore, cannot be used to continuously monitor CO. Real-time PCMR can be used for this application, but ensuring adequate spatiotemporal resolution requires significant data undersampling, often combined with efficient trajectories (e.g. spiral)[8].

Unfortunately, reconstruction of highly accelerated non-Cartesian data is time-consuming due to the iterative nature of state-of-the art reconstruction algorithms (e.g. compressed sensing (CS)). This problem has been partly mitigated by graphics processing units (GPU) that enable low latency reconstruction of continuously acquired real-time data [9]. However, processing the large amounts of data produced is time consuming and current methods do not enable real-time monitoring.

We have recently shown that deep learning (DL) can be used to remove aliasing artefact (deep artefact suppression) from both magnitude and phase images of highly accelerated real-time spiral PCMR data acquired at rest [10]. The U-Net architecture used for deep artefact suppression was originally proposed for segmentation of biomedical images [11] and performed excellently for a wide range of segmentation applications [12]. We propose to extend our previous work [10] by performing both low latency deep artifact suppression and segmentation on real-time flow data and combining this with optimized communication and visualization for near real-time monitoring of CO during exercise. The specific aims were: i) to develop and demonstrate the feasibility of a low-latency continuous CO monitoring framework on the scanner, in 10 healthy subjects during a simple exercise study, ii) compare CO measurements at rest between our proposed "Flow Reconstruction and Segmentation for low latency CO monitoring" (FReSCO) method and reference free-breathing retrospectively ECG-gated Cartesian PCMR and ii) compare CO measurements at rest and peak exercise between FReSCO and CS reconstruction.

## Methods

Our proposed framework for low-latency CO monitoring relies on: i) a highly accelerated real-time spiral PCMR acquisition, ii) an open-source cross-platform communication framework (Gadgetron), and iii) two sequential U-Nets for fast deep artefact suppression and segmentation of the real-time data. The FReSCO framework is illustrated in Figure 1.

*Acquisition:*

The real-time flow acquisition uses a golden angle variable density spiral [13] trajectory, with the outer 10% of k-space 2.5x less densely sampled than the inner 20% (with linearly decreasing density in-between - trajectory depicted in Supporting Information Figure S1). One sided velocity encoding was achieved by acquiring each readout twice (velocity encoded and compensated) and three spiral interleave positions (6 readouts) were acquired per frame leading to an acceleration factor of ~8.7/21.7 for inner/outer k-space. Scan parameters included - FOV: 400×400mm, voxel-size: 2.1×2.1×6.0mm, TR/TE: 5.8/2.1ms, temporal resolution: 35.0ms, flip angle: 20º, VENC: 200cm/s.

Reference standard flow imaging was acquired for comparison at rest using a free-breathing, retrospectively ECG-gated, Cartesian PCMR sequence with the following parameters - FOV: 254×370mm, voxel-size: 1.4×1.4×6.0mm, TR/TE: 5.1/2.7ms, temporal resolution: ~30ms, flip angle: 20º, VENC: 180cm/s, averages: 3, acquisition time: 95.71 ± 21.4s.

*Training Data:*

The FReSCO DL framework consisted of independently trained deep artefact suppression and aortic segmentation networks. The training data for deep artefact suppression was created from 516 breath-hold, retrospectively ECG-gated, uniform density spiral PCMR [8] datasets in the aortic position of patients with pediatric and/or congenital heart disease (age: 21±13.5yrs, heart rate: 71±12 beats per minute (bpm)). Each dataset consisted of magnitude and phase subtracted images (as stored within clinical routine). To create the paired synthetically corrupted and truth images for training, the complex data was first Fourier transformed and undersampled with the proposed trajectory. The synthetic undersampled k-space data was then inverse Fourier transformed to produce the aliased images. This deep artifact suppression dataset was split into 470/30/16 for training, validation and testing.

The training data for segmentation was created from 96 of the above datasets using a semi-automatic method based on an optical flow registration with manual operator correction [14] by an expert (V.M.). This segmentation dataset (complex images as input and segmentation masks as output) was split into

70/10/16 for training, validation and testing (with the same test set as used for the artefact suppression network, above).

Collection of all retrospective data was approved by the national research ethics committee (Ref. 06/Q0508/124).

*Networks and Training:*

The overall DL framework for deep artefact suppression and segmentation is shown in Figure 2.A and consisted of two consecutive, independently trained, 3D U-Nets. The artefact suppression model was trained on center cropped (128x128), paired complex corrupted and truth images in blocks of 24 frames, with real and imaginary channels, using an average 2D structural similarity index (AvgSSIM) based loss ($L$) [10]:

$$L = 1 - \frac{SSIM\left(\frac{real(y)+1}{2}, \frac{real(\hat{y})+1}{2}\right) + SSIM\left(\frac{imag(y)+1}{2}, \frac{imag(\hat{y})+1}{2}\right)}{2}$$

Where $y$ is the ground truth image, $\hat{y}$ is the predicted image, and $real$ and $imag$ are the real and imaginary components. Data augmentation was performed during training, and included random smooth phase offsets, image flips, rotations and roll (i.e. time shift). In addition, translational motion was applied to 50% of cases to simulate exercise and rest cases. A Hyperband optimization [15] was performed to choose the optimum U-Net parameters, including: number of scales, initial filters, convolution blocks per scale and learning rate (the tunable U-Net architecture is depicted in Supporting Information Figure S2 and range of parameters explored in Supporting Information Table 1).

The segmentation model was also trained on 24 frames of cropped (128x128) paired complex images (as real and imaginary channels) and segmentation maps. The magnitude data underwent contrast limited adaptive histogram equalization (CLAHE) [16] for more robust segmentation. Data augmentation and hyperparameter optimization was broadly the same as in the artefact suppression model, except that the type of segmentation loss was also optimized (as described in Supporting Information Table 1).

Hyperband parameters for the deep artifact suppression/segmentation models included: 100/150 maximum number of epochs, hyperband factor of 3 (discarded proportion within each bracket) and 172/172 configurations tested. All training was performed using TensorFlow [17] on a Linux Workstation (with NVIDIA TITAN RTX 24GB).

*In-silico Validation:*

Evaluation of both tasks were performed on the test set (including motion in all 16 cases) using: 1) imaging metrics: Mean Absolute error (MAE), Peak Signal to Noise Ratio (PSNR) and AvgSSIM; 2) and segmentation metrics: Binary Cross Entropy (BCE) and Dice score. The segmentation network's performance was tested on both the original 'truth' test images and on the DL restored corrupted images to evaluate any loss in performance due to the DL reconstruction.

*Prospective Experiments:*

10 healthy subjects (age: 33.2±4.3yrs) were prospectively acquired (Aera 1.5T, Siemens Healthineers). Reference standard resting aortic PCMR data was acquired using a retrospectively ECG-gated, breath-hold cartesian PCMR scan followed by three minutes of continuous real-time imaging (5143 frames) during rest, exercise and recovery.

A simple exercise protocol was used to demonstrate feasibility of the framework consisting of 40 seconds of rest, 80 seconds of moderate exercise and 60 seconds of recovery. The exercise consisted of supine repeated leg extensions (following a metronome at 1 beat per second) using an extensible band held by the subject to provide resistance.

Collection of prospective data was approved by the national research ethics committee (Ref. 17/LO/1499), and written consent was obtained in all subjects.

*In-vivo Real-time Reconstruction:*

Prospective images were reconstructed in near real-time during scanning using Gadgetron [18] for low latency communication with an external reconstruction and visualization computer (Linux Workstation with NVIDIA GeForce RTX 3060 12GB). TensorFlow MRI [19] was used for gridding and the proposed networks were used for deep artifact suppression and segmentation.

At the start of acquisition, the framework was initialized. This included setting up the Gadgetron pipeline, importing the necessary modules, and computing trajectories, density compensation weights [20], coil sensitivity maps [21] (from first 10 frames of data) and Maxwell correction terms [22].

Figure 2 depicts the pipeline at inference. Flow encoded and flow compensated frames were gridded, coil combined, cropped, normalized and buffered into 2D+time blocks. Deep artefact suppression was then performed in a sliding window fashion (window = 24 frames, step size = 18 frames, keeping only central frames to remove edge effects) separately on the flow encoded and compensated blocks.

The artifact suppressed data were then combined (average magnitude and phase subtraction), and the magnitude images were equalized using CLAHE. The magnitude and phase subtracted data were then passed to the segmentation network (as real and imaginary channels) for aortic segmentation. This segmented data was used to quantify flow from the phase subtracted PCMR data (after Maxwell correction). Peak detection was then performed for real-time monitoring of heart rate, stroke volume and cardiac output. The resultant images, segmentation and flow curves were displayed with low latency on the external computer (interface shown in Figure 1.B).

Reconstruction timings were recorded during acquisition on the last 1000 frames. Latency is approximated as the time between the start of an acquisition block of 24 images, until completion of the reconstruction and segmentation of the block of images.

Real-time images at rest (seconds 4 to 14) and at peak exercise (seconds 107 to 117) were additionally reconstructed offline to compare the proposed method with a CS reconstruction of the same data using temporal total variation regularization (BART [23] toolbox). The CS regularization factor was set empirically ($\lambda$=5E-4) and data was reconstructed in the same blocks as the DL reconstruction (window = 24 frames, step size = 18 frames, keeping only central frames). Cardiac output was extracted for both proposed DL and CS methods using the same segmentation (obtained from proposed DL segmentation network of the DL reconstructed images) to limit the source of discrepancies to reconstruction differences only.

Additionally, real-time CO obtained at rest was compared to the reference standard PCMR, which was segmented using a semi-automatic method based on an optical flow registration with manual operator correction (by an expert J.B.).

*Statistical Analysis:*

Statistical analyses were performed using Python. All compared distributions were tested for normality using Shapiro-Wilk tests. In-silico metrics were compared using paired t-tests. In-vivo, Bland-Altman analysis of the prospective extracted flow volumes was performed between the reference PCMR (at rest only), FReSCO and CS measurements. Cardiac output mean biases and limits of agreements (LOA) were reported. These biases were tested for statistical significance using a repeated measurements one-way ANOVA (if >2 groups) or paired t-tests otherwise.

# Results

*Training and In-silico Validation:*

Training with hyperparameter optimization of the deep artifact suppression and segmentation networks took 67 and 10 hours respectively. The range of parameters explored and final selected network parameter values for both tasks are shown in Supporting Information Table S1.

In-silico results are shown for four representative test subjects in Supporting Information Figure S3 (and one representative subject in Supporting Information Video S1). For deep artifact suppression the MAE, PSNR and AvgSSIM were 0.023±0.004, 29.3±1.4 and 0.88±0.03 respectively. For the segmentation model the BCE and Dice were 0.048±0.054 and 0.87±0.13. It should be noted that segmentation accuracy on restored images was not statistically significantly different from the segmentation accuracy obtained from 'truth' images (BCE 0.061, p=0.09; Dice 0.87, p=0.71).

*Feasibility of proposed method:*

A video of the proposed interface for real-time monitoring as recorded during scanning is provided in Supporting Information Video S2 and a snapshot of the interface at the end of acquisition in Figure 1.B. FReSCO was able to adequately remove artefact and segment the aorta with a latency <1s both at rest and during exercise without any user interaction. The CO increased from 5.82±1.10L/min at rest to 7.42±1.34L/min at peak exercise and HR increased from 68±8bpm at rest to 94±8bpm at peak exercise. Two representative curves recorded during exercise are shown in Figure 3 and demonstrate different responses to exercise better seen with continuous monitoring.

A schematic of reconstruction timings is shown in Figure 2.B. The gridding time was ~16.2ms/frame compared to an acquisition time of ~35ms/frame. After the final frame in a block was gridded, deep artifact suppression of both encodings and segmentation of a block took on average 151ms. It should be noted that initialization of the pipeline led to an initial latency of about 16s. However, as the total reconstruction time for a block was shorter than the acquisition time (for 18 central frames, gridding and deep artifact suppression added up to ~443ms vs. ~630ms of acquisition), the reconstruction was able to catch up after 26 seconds (42 seconds into the acquisition). After this transition period, mean latency was about 622ms for the central frame of the block, and 902ms for the first frame.

*Comparison with CS reconstruction and reference standard PCMR:*

Magnitude and phase subtracted images, reconstructed using FReSCO and CS, as well as reference ECG-gated data, are shown in Figure 4 (Supporting Information Video S3). Time-averaged and real-time curves are shown for one subject in Supporting Information Figure S4. At rest, there was good agreement between the proposed method and the reference (Figure 5.A) with no-significant differences in CO (Bias = -0.21±0.50 L/min, p=0.246). There was a small but statistically significant negative bias in CO between CS and the reference (Figure 5.B, Bias=-0.38±0.35L/min, p=0.009) and a trend towards significance between proposed DL method and CS (Figure 5.C, Bias=0.18±0.24L/min, p=0.052). During exercise, there was good agreement in peak CO between CS and FReSCO (Figure 5.D, Bias=0.12±0.48 L/min, p=0.458).

# Discussion

In this study, we demonstrated the feasibility of low latency real-time cardiac output monitoring using real-time PCMR, combined with DL-based artifact suppression and automated segmentation during exercise. There was good agreement between CO measured using this method and conventional gated PCMR (at rest), and CS reconstruction of the same data (at rest and exercise). Thus, we believe our approach has the potential to simplify continuous CO measurement during various stress protocols.

*Networks and training:*

We have previously shown the utility of deep artifact suppression for fast reconstruction of highly undersampled spiral real-time PCMR [10]. As our application also required processing of PCMR data, we additionally trained a network for fully automated segmentation. We showed excellent network performance for both tasks, and the combined inference time was short enough for a latency <1s. Thus, more sophisticated joint reconstruction and segmentation networks [24,25] that could have had shorter inference times were not necessary. Further benefits of splitting the tasks were that it allowed the use of a larger dataset for learning deep artifact suppression, different preprocessing of the inputs (e.g. CLAHE prior to segmentation), and separate optimization of network hyperparameters. Future improvements could include using multi-coil raw data for training [26], and using a more end-to-end approach to optimize acquisition, reconstruction [27–29], and segmentation [24].

*Real-time reconstruction and visualization:*

The proof-of-concept framework was demonstrated at the scanner during continuous exercise. It was able to remove artifacts and segment images without user intervention and with low enough latency to provide almost real-time monitoring. Latency could be further reduced by including: 1) initialization of the pipeline prior to the start of acquisition to reduce the initial latency, 2) parallelization of gridding and deep artifact suppression for higher frame rate, and 3) using a memory-based network to reconstruct the latest frame rather than blocks, while still using temporal redundancies [30].

*Comparison with CS reconstruction and gated reference standard:*

There was reasonable agreement between FReSCO and the reference standard ECG-gated PCMR sequence for measurement of CO at rest. There was also reasonable agreement between the proposed DL and CS reconstructions of the same data at rest and exercise. This is in keeping with previous larger studies comparing deep artifact suppression to CS and gated PCMR. However, in our case agreement with reference PCMR also relied on the accuracy of automated segmentation. The good agreement suggests that both deep artifact suppression and automated segmentation worked robustly within our framework. However, further testing in larger and more heterogenous populations is required prior to more general use.

*Applications:*

Stress testing, particularly with exercise, is becoming increasingly important in cardiac MRI as it provides important information about hemodynamic responsiveness. However, it has been difficult to continuously monitor cardiac output due to long reconstruction times and difficulty in segmenting thousands of frames of PCMR data. We have demonstrated this is easily achieved with our framework, and we showed that the dynamic response to exercise was highly variable. Thus, continuous CO data may provide new insights into cardiovascular disease. In future work we aim to assess the feasibility of this approach in more strenuous forms of in-scanner exercise (e.g. recumbent bicycle).

# Conclusion

FReSCO enables real-time monitoring of CO during exercise and could greatly simplify workflow and provide a convenient tool for assessment of the hemodynamic response to a range of stressors (e.g. exercise, adenosine, dobutamine, eating, and mental tasks). Future work will aim to generalize the framework to multiple vessels of interest and utilize the proposed framework within research protocols.

# Acknowledgements

This work was supported by the British Heart Foundation (grant: NH/18/1/33511) and UKRI FLF (grant: MR/S032290/1). This work used the open-source framework Gadgetron (https://github.com/gadgetron/gadgetron) and TensorFlow MRI (https://github.com/mrphys/tensorflow-mri).

Figures

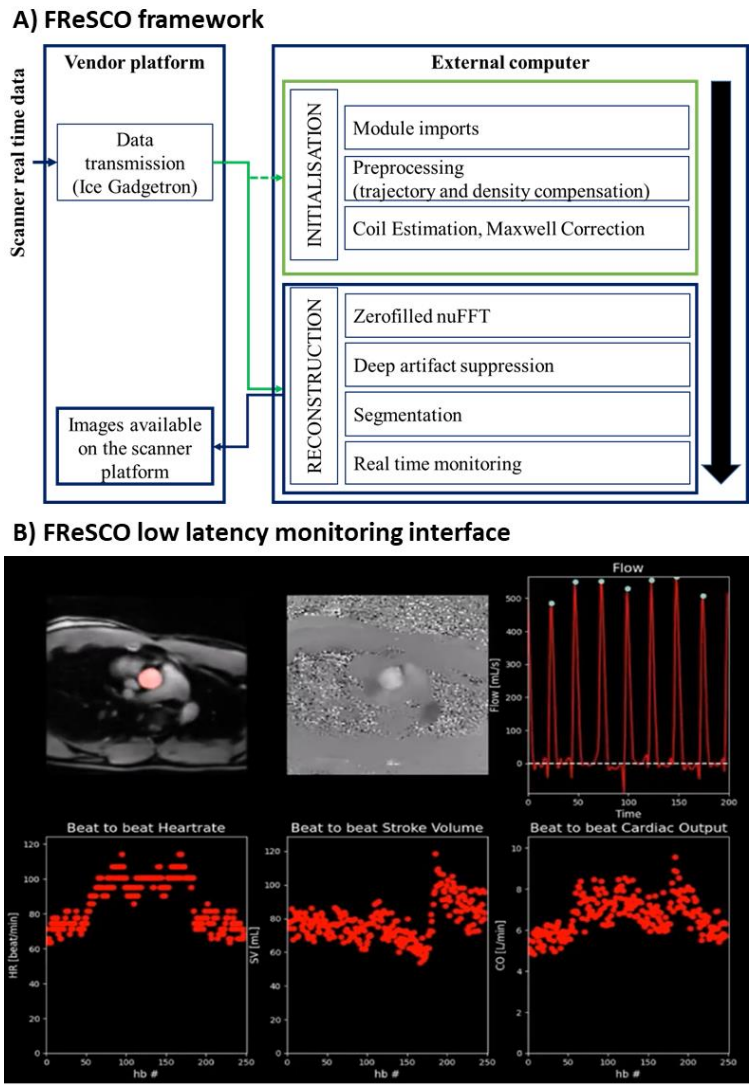

**Figure 1. FReSCO framework for low latency monitoring of aortic flow.** A) Real-time golden angle variable density spiral data is forwarded to an external computer using Gadgetron. The pipeline is initialized using the 10 first frames (i.e. Module imports, trajectory, density compensation, coil estimation, Maxwell correction) and the proposed reconstruction and flow monitoring are performed during scanning. Flow maps and segmentations are sent back to the scanner. B) Illustration of the proposed real-time monitoring interface at the end of a three minute exercise scan. Top-left: Segmented magnitude images, top-middle: phase images, top-right: flow curves, bottom-left: beat-to-beat heartrate, bottom-middle: stroke volume and bottom-right: cardiac output.

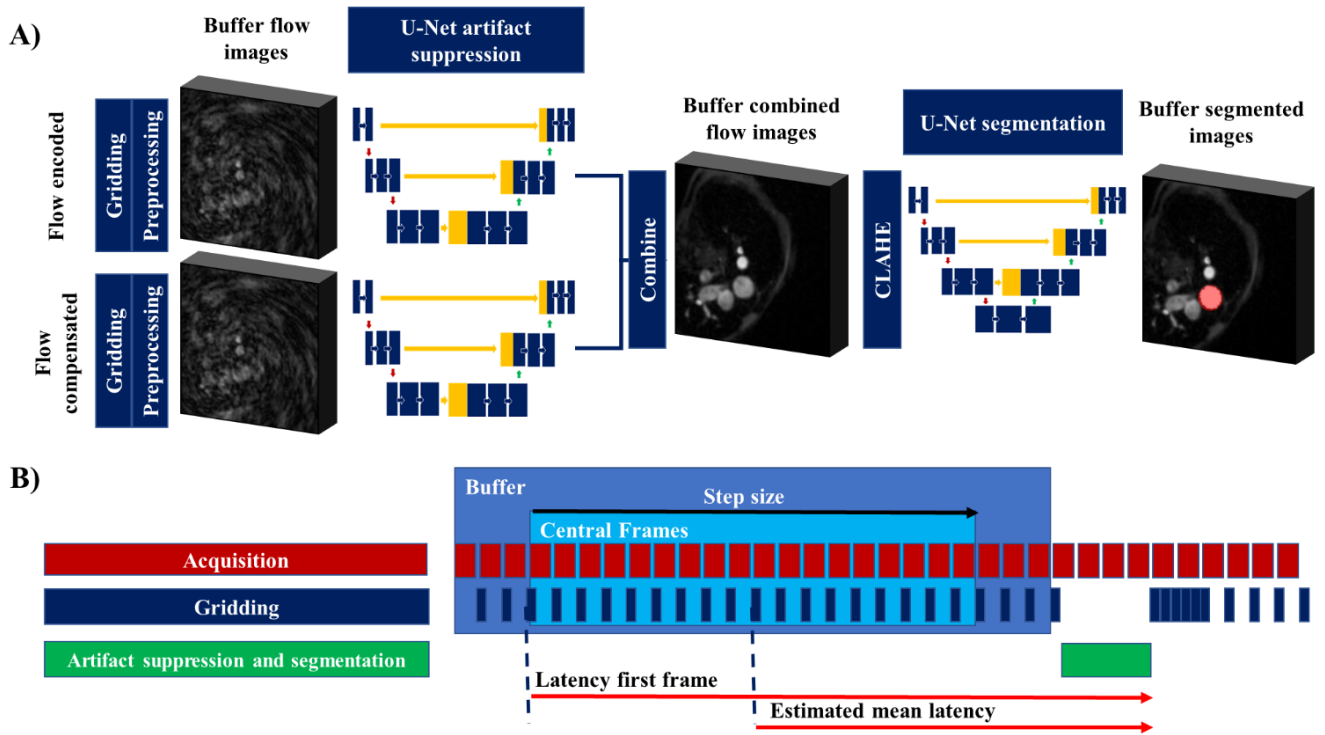

**Figure 2.** A) Overview of the FReSCO framework at inference. Deep artifact suppression is performed on the buffered gridded flow encoded and compensated data. Segmentation is performed on the combined, magnitude contrast limited adaptive histogram equalized (CLAHE) images. B) Sliding window reconstruction timings (at scale). Each acquired frame is gridded independently but deep artifact suppressed and segmented as a block of 24 frames.

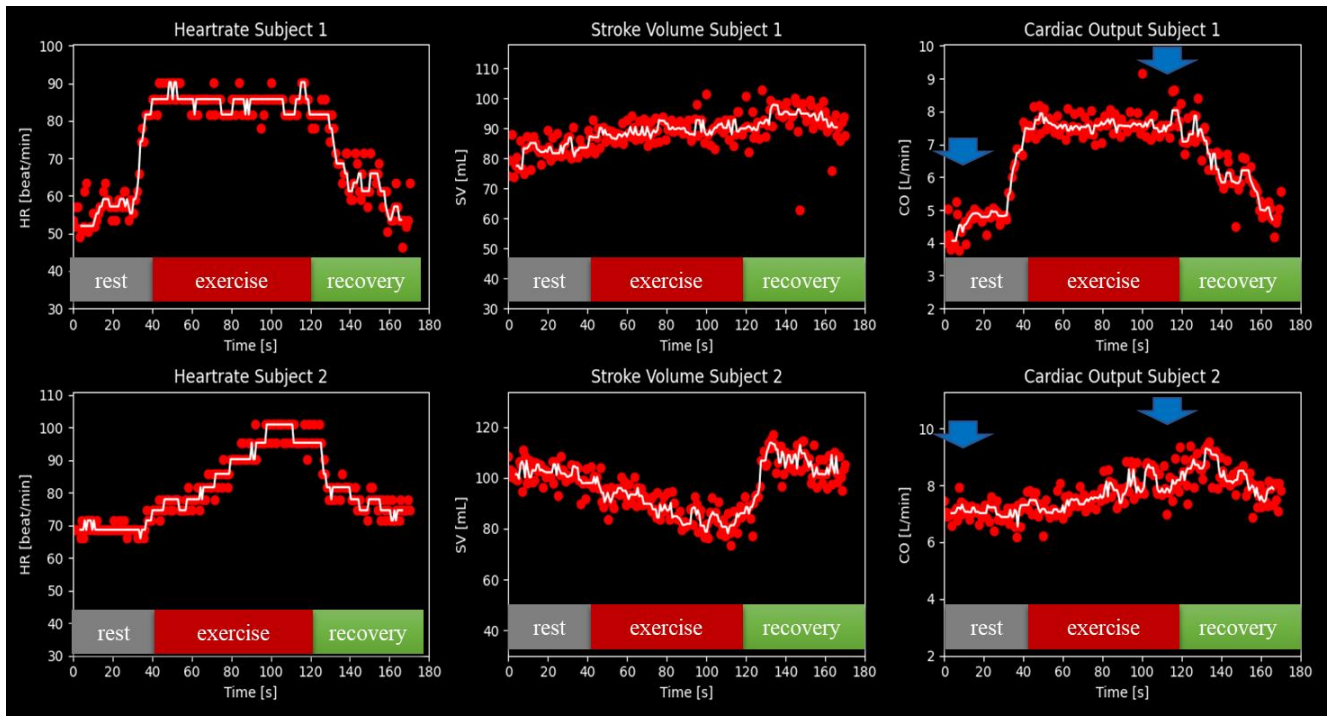

**Figure 3.** Heartrate, Stroke Volume and Cardiac output curves obtained from two subjects with different responses to exercise. In white, the median filtered curves. Blue arrows depict the 10 seconds areas used for comparison to CS and reference at rest (seconds 4 to 14) and CS only at peak exercise (seconds 107 to 117).

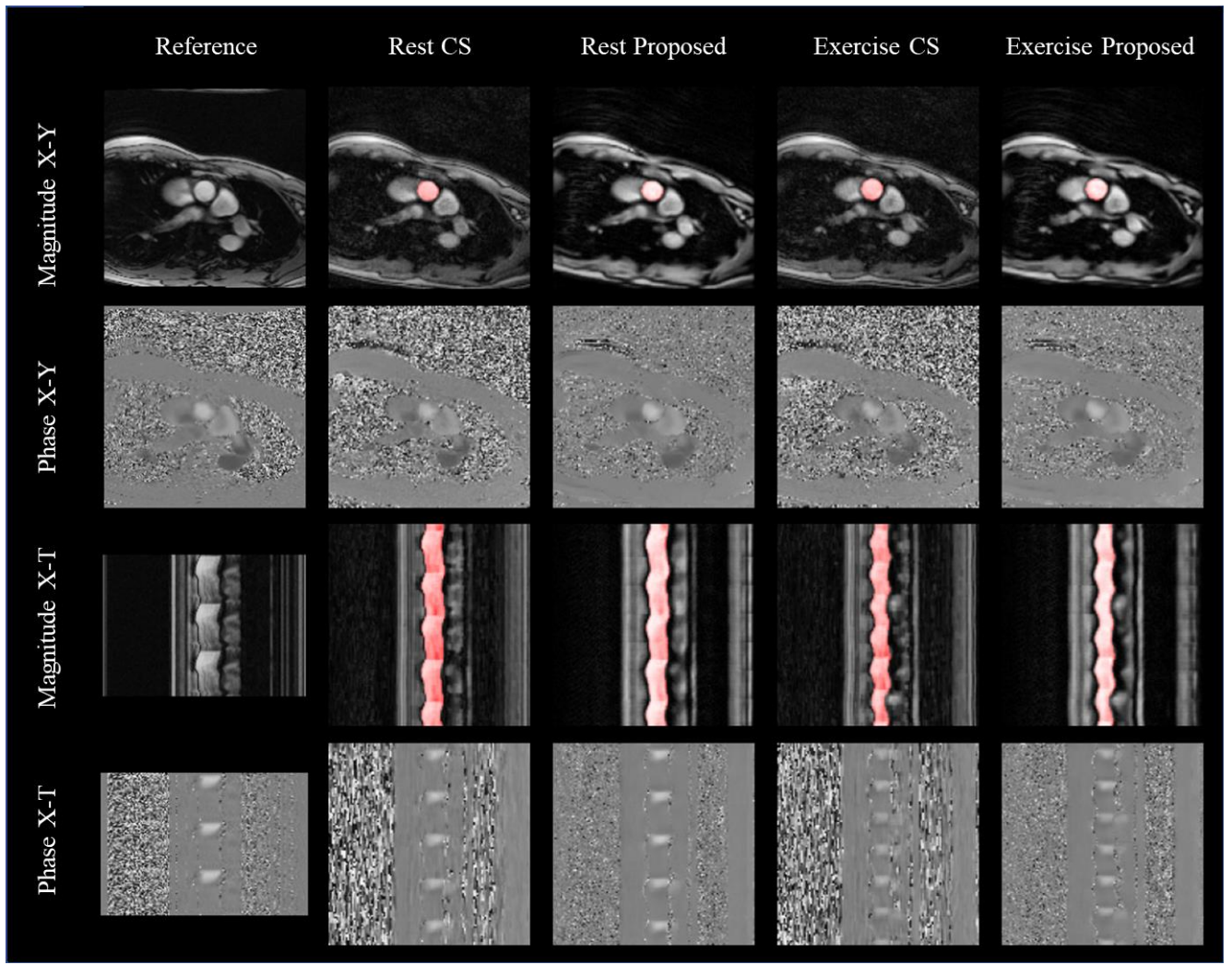

**Figure 4.** Reference (repeated for three cycles), Compressed sensing (CS) and FReSCO images at rest, and CS and FReSCO images during exercise in a representative subject. Magnitude x-y, magnitude x-t, phase x-y and phase x-t images are shown. Automatic segmentations (computed from the proposed DL images) are overlaid on top of the magnitude images. Corresponding video in Supporting Information Video S3.

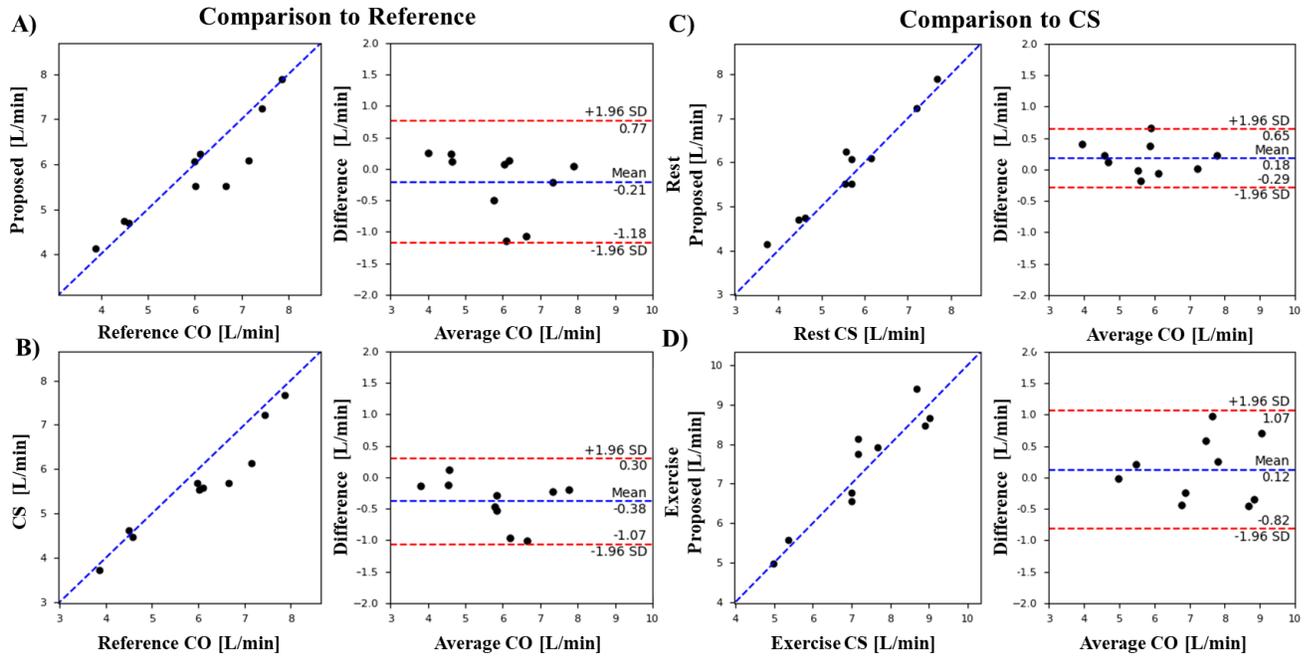

**Figure 5.** Correlation and Bland Altman plots comparing Cardiac Output (CO) from A) Proposed FReSCO to reference at rest, B) Compressed sensing (CS) to reference at rest, C) FReSCO to CS at rest and D) FReSCO to CS at peak exercise.

Supporting Figures

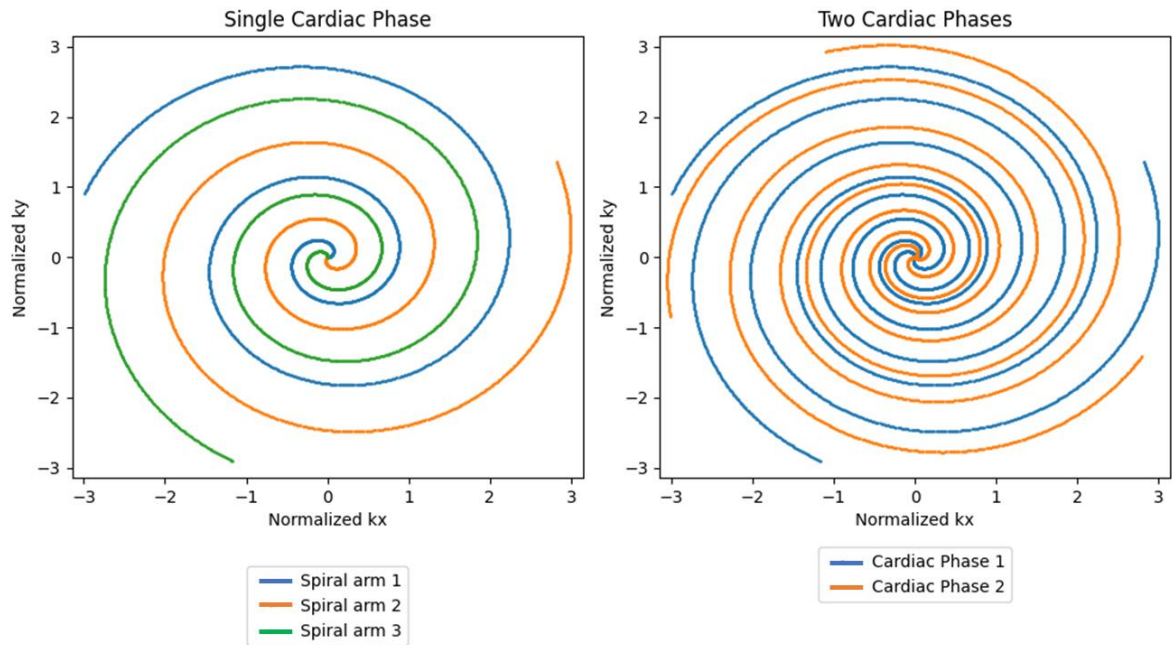

**Supporting Information Figure S1.** Left: Example of variable density golden angle spiral trajectory for one cardiac phase image. Three spiral arms rotated by the golden angle are combined for each cardiac phase with a golden angle increment. The same spiral arm is acquired twice consecutively with and without flow encoding. Each spiral arm is accelerated by x26 in the center 20% of k-space and x65 in the outermost 20% of k-space. Right: Trajectories covered in two consecutive cardiac phases. Each frame is accelerated by x8.7 in the center 20% of k-space and x21.7 in the outermost 20% of k-space.

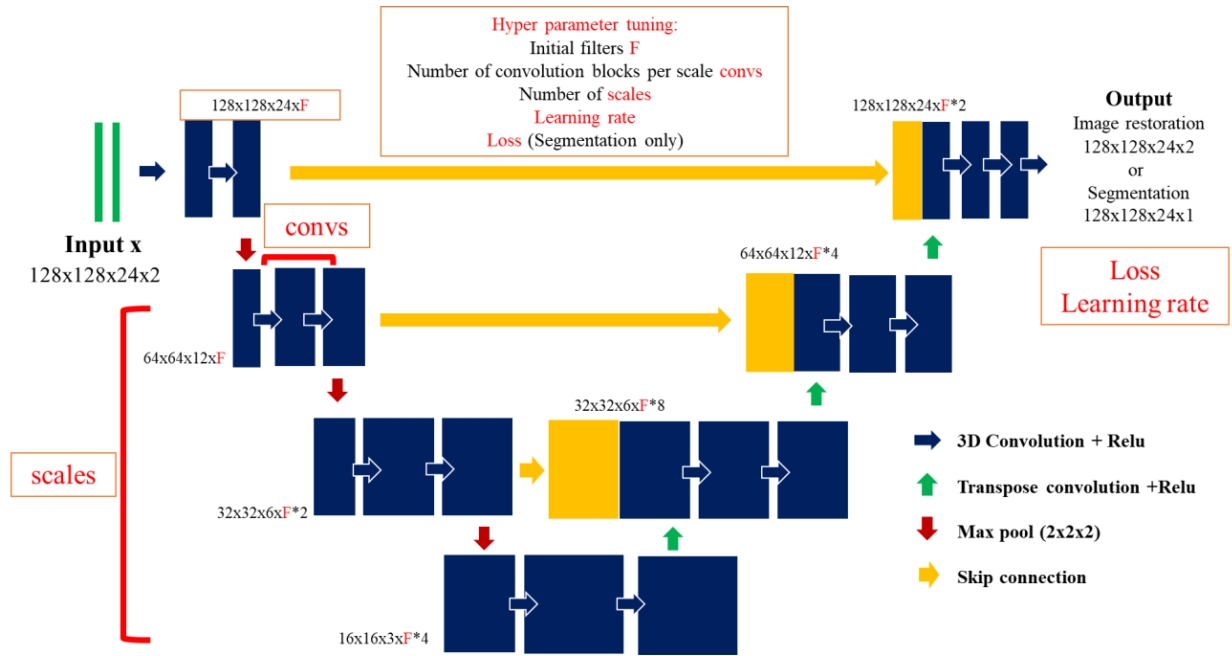

**Supporting Information Figure S2.** Diagram showing the U-Net architecture and explored hyper parameters (in red). Range explored and final selected parameter values for both tasks are shown in Supporting Information Table S1.

| A) Training | Flip Rotation Roll | Motion probability | Apply | Crop | Normalization | Hyperband Max Epochs |
|---|---|---|---|---|---|---|
| Deep artifact suppression | Images | Train: 0.5 Val: 0.5 Test: 1 | Phase offset Undersampling | 128x128 | Scaled Magnitude [0:1] | 100 |
| Segmentation | Images and segmentations | Train: 0.5 Val: 0.5 Test: 1 | CLAHE | 128x128 | Scaled Magnitude [0:1] | 150 |
| **B) Tuned Parameters** | **Scales** | **Blocks** | **Initial filters** | **Loss** | **Learning rate** | **Objective** |
| Deep artifact suppression | 2 [1:3] | 2 [1:3] | 19 [16:32] | AvgSSIM N/A | 4,64E-4 [1E-3:1E-5] | Validation AvgSSIM |
| Segmentation | 3 [1:3] | 1 [1:3] | 28 [16:32] | Focal tversky [Dice, Jaccard, Focal tversky, BCE] | 5.41E-4 [1E-3:1E-5] | Validation Dice |

**Supporting Information Table S1.** A) Information relative to the data augmentation and training of individual networks. B) Resulting values selected from the hyperband optimization as well as the range explored for each hyper-parameter.

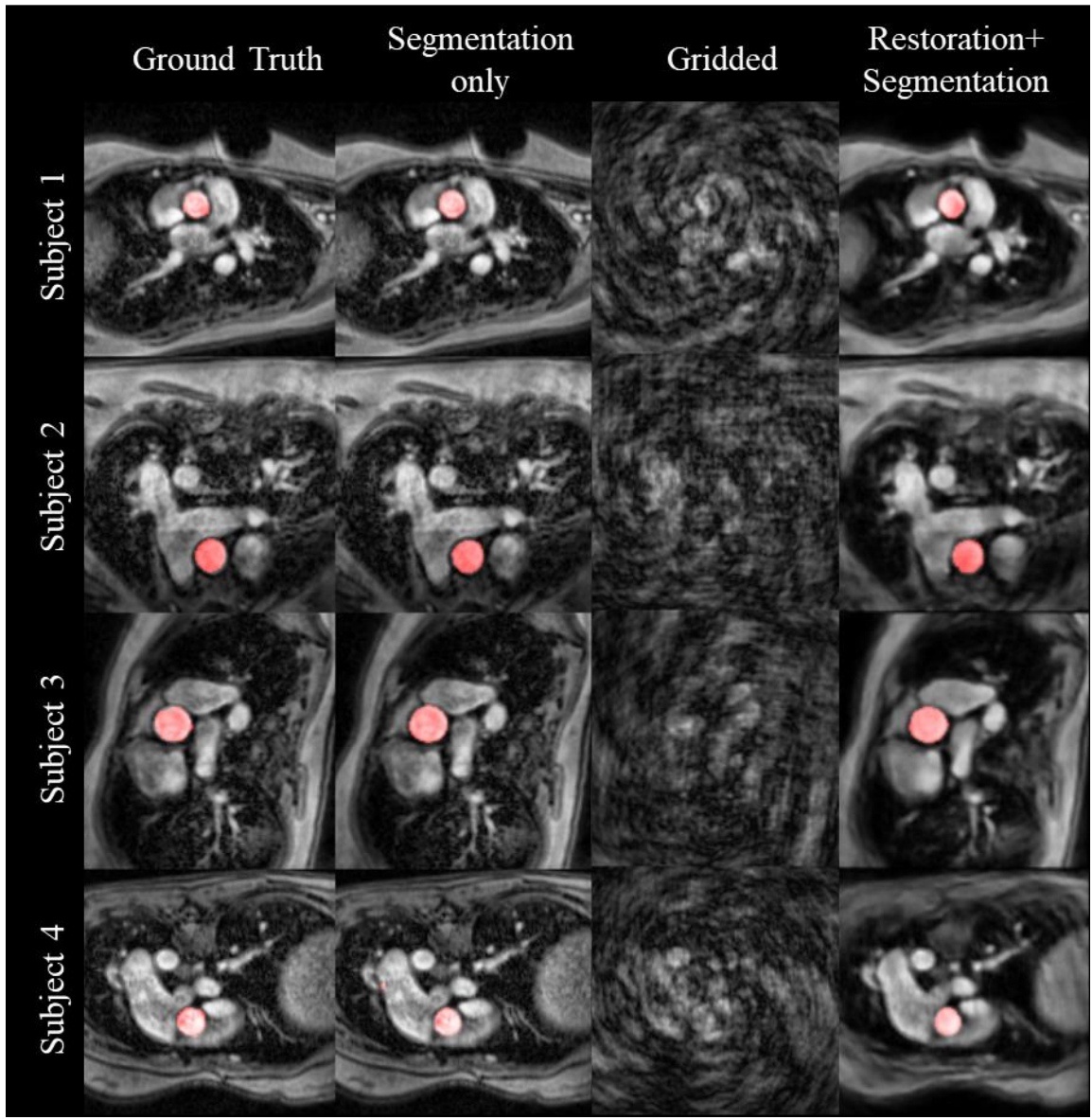

**Supporting information Figure S3.** Top to Bottom: Four representative test subjects. Left to right: Ground Truth images and segmentation, Ground truth images and DL segmentation calculated from Ground Truth images, undersampled images (input to deep artifact suppression network) and DL restored images and DL segmentation (estimated from DL images). The segmentations are overlaid in red when applicable.

### A) Rest
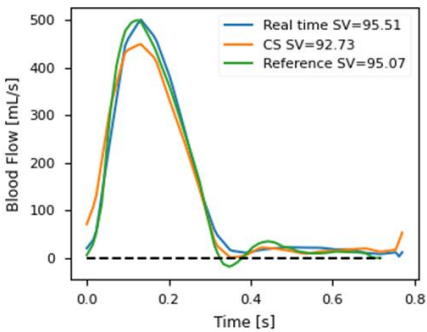
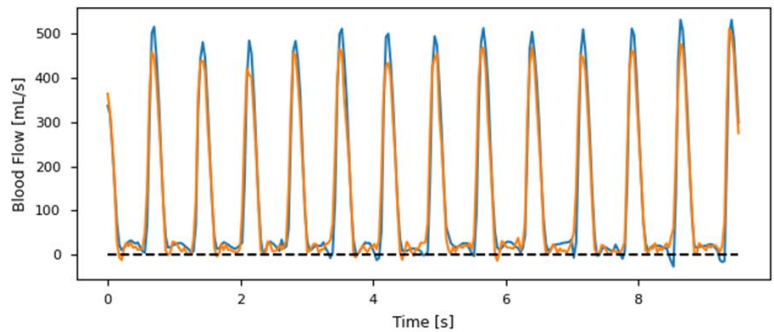

### B) Exercise
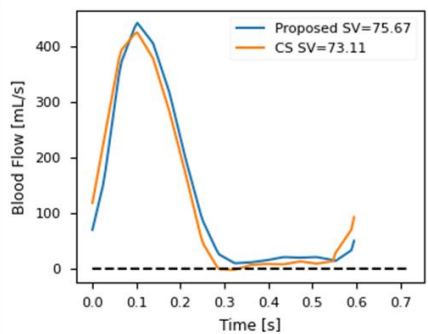
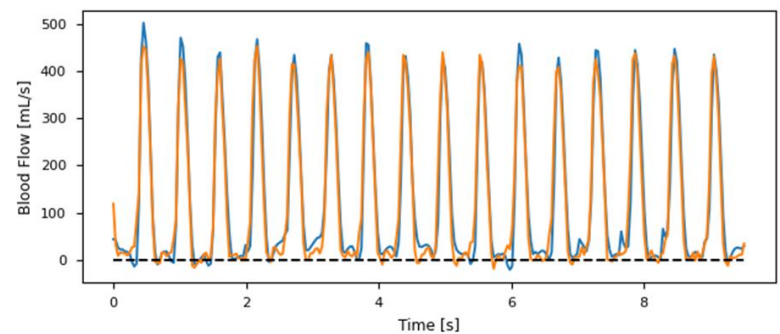

**Supporting Information Figure S4.** Left: Averaged FReSCO, averaged CS and reference flow curves and Right: real-time FReSCO and CS flow curves obtained A) at rest and B) during exercise showing good agreement between methods.

**Supporting Information Video S1.** Representative test set subject Video. Top row: Magnitude ground truth images and segmentation, overlaid predicted segmentation from ground truth, undersampled input, restored images and overlaid predicted segmentation from restored images. Bottom row: Matching phase images.

**Supporting Information Video S2. Real-Time flow monitoring during exercise.** Video of the interface during the start of exercise, peak exercise and end of recovery (seconds 40-50, 110-120 and 170-180) Top Row: Magnitude and overlaid segmentation, Phase and extracted blood flow curve with marked detected peaks. Bottom Row: Beat-to-beat heartrate, stroke volume and cardiac output as provided in real-time.

**Supporting Information Video S3.** Comparison video depicting magnitude, segmentations and flow maps for Reference, CS at rest, FReSCO at rest, CS at exercise and FReSCO at exercise of the same subject.

# Bibliography

[1] Zhang S, Joseph AA, Voit D, Schaetz S, Merboldt K-D, Unterberg-Buchwald C, et al. Real-time magnetic resonance imaging of cardiac function and flow—recent progress. Quant Imaging Med Surg 2014;4:313. https://doi.org/10.3978/J.ISSN.2223-4292.2014.06.03.

[2] Joseph AA, Voit D, Frahm J. Inferior vena cava revisited - Real-time flow MRI of respiratory maneuvers. NMR Biomed 2020;33. https://doi.org/10.1002/NBM.4232.

[3] Sun A, Zhao B, Li Y, He Q, Li R, Yuan C. Real-time phase-contrast flow cardiovascular magnetic resonance with low-rank modeling and parallel imaging. J Cardiovasc Magn Reson 2017;19:1–13. https://doi.org/10.1186/S12968-017-0330-1/FIGURES/9.

[4] Joseph A, Kowallick JT, Merboldt KD, Voit D, Schaetz S, Zhang S, et al. Real-time flow MRI of the aorta at a resolution of 40 msec. J Magn Reson Imaging 2014;40:206–13. https://doi.org/10.1002/JMRI.24328.

[5] White SW, Quail AW, De Leeuw PW, Traugott FM, Brown WJ, Porges WL, et al. Impedance cardiography for cardiac output measurement: An evaluation of accuracy and limitations. Eur Heart J 1990;11:79–92. https://doi.org/10.1093/EURHEARTJ/11.SUPPL_I.79.

[6] Hecht HS, DeBord L, Sotomayor N, Shaw R, Dunlap R, Ryan C. Supine Bicycle Stress Echocardiography: Peak Exercise Imaging is Superior to Postexercise Imaging. J Am Soc Echocardiogr 1993;6:265–71. https://doi.org/10.1016/S0894-7317(14)80062-X.

[7] Klein C, Schalla S, Schnackenburg B, Bornstedt A, Fleck E, Nagel E. Magnetic resonance flow measurements in real time: Comparison with a standard gradient-echo technique. J Magn Reson Imaging 2001;14:306–10. https://doi.org/10.1002/JMRI.1187.

[8] Steeden JA, Atkinson D, Hansen MS, Taylor AM, Muthurangu V. Rapid flow assessment of congenital heart disease with high-spatiotemporal- resolution gated spiral phase-contrast MR imaging. Radiology 2011;260:79–87. https://doi.org/10.1148/radiol.11101844.

[9] Kowalik GT, Steeden JA, Pandya B, Odille F, Atkinson D, Taylor A, et al. Real-time flow with fast GPU reconstruction for continuous assessment of cardiac output. J Magn Reson Imaging 2012;36:1477–82. https://doi.org/10.1002/jmri.23736.

[10] Jaubert O, Steeden J, Montalt-Tordera J, Arridge S, Kowalik GT, Muthurangu V. Deep artifact suppression for spiral real-time phase contrast cardiac magnetic resonance imaging in congenital

heart disease. Magn Reson Imaging 2021;83:125–32. https://doi.org/10.1016/J.MRI.2021.08.005.

[11]  Ronneberger O, Fischer P, Brox T. U-net: Convolutional networks for biomedical image segmentation. Lect. Notes Comput. Sci. (including Subser. Lect. Notes Artif. Intell. Lect. Notes Bioinformatics), vol. 9351, Springer Verlag; 2015, p. 234–41. https://doi.org/10.1007/978-3-319-24574-4_28.

[12]  Isensee F, Petersen J, Klein A, Zimmerer D, Jaeger PF, Kohl S, et al. nnU-Net: Self-adapting Framework for U-Net-Based Medical Image Segmentation. Inform Aktuell 2018:22. https://doi.org/10.1007/978-3-658-25326-4_7.

[13]  Pipe JG, Zwart NR. Spiral trajectory design: A flexible numerical algorithm and base analytical equations. Magn Reson Med 2014;71:278–85. https://doi.org/10.1002/MRM.24675.

[14]  Odille F, Steeden JA, Muthurangu V, Atkinson D. Automatic segmentation propagation of the aorta in real-time phase contrast MRI using nonrigid registration. J Magn Reson Imaging 2011;33:232–8. https://doi.org/10.1002/jmri.22402.

[15]  Li L, Jamieson K, DeSalvo G, Rostamizadeh A, Talwalkar A. Hyperband: A Novel Bandit-Based Approach to Hyperparameter Optimization. J Mach Learn Res 2016;18:1–52.

[16]  Pizer SM, Amburn EP, Austin JD, Cromartie R, Geselowitz A, Greer T, et al. Adaptive histogram equalization and its variations. Comput Vision, Graph Image Process 1987;39:355–68. https://doi.org/10.1016/S0734-189X(87)80186-X.

[17]  Abadi M, Barham P, Chen J, Chen Z, Davis A, Dean J, et al. TensorFlow: A System for Large-Scale Machine Learning TensorFlow: A system for large-scale machine learning. Proc 12th USENIX Symp Oper Syst Des Implement 2016:265–83.

[18]  Hansen MS, Sørensen TS. Gadgetron: An open source framework for medical image reconstruction. Magn Reson Med 2013;69:1768–76. https://doi.org/10.1002/mrm.24389.

[19]  Montalt-Tordera J, Steeden J, Muthurangu V. TensorFlow MRI: A Library for Modern Computational MRI on Heterogenous Systems. Proc ISMRM London 2022 2022.

[20]  Pipe JG, Menon P. Sampling Density Compensation in MRI: Rationale and an Iterative Numerical Solution. Magn Reson Med 1999;41:179–86. https://doi.org/10.1002/(SICI)1522-2594(199901)41:1.

[21]  Walsh DO, Gmitro AF, Marcellin MW. Adaptive Reconstruction of Phased Array MR Imagery

2000. https://doi.org/10.1002/(SICI)1522-2594(200005)43:5.

[22]  Bernstein MA, Zhou XJ, Polzin JA, King KF, Ganin A, Pelc NJ, et al. Concomitant gradient terms in phase contrast MR: analysis and correction. Magn Reson Med 1998;39:300–8. https://doi.org/10.1002/MRM.1910390218.

[23]  Uecker M, Ong F, Tamir JI, Bahri D, Virtue P, Cheng J, et al. Berkeley Advanced Reconstruction Toolbox. Annu Meet ISMRM, Toronto 2015 2015;23:2486.

[24]  Buchholz T-O, Prakash M, Krull A, Jug F. DenoiSeg: Joint Denoising and Segmentation. ArXiv 2020:2005.02987v2.

[25]  Jaubert O, Montalt-Tordera J, Arridge S, Steeden JA, Muthurangu V. Joint deep artifact suppression and segmentation network for low latency beat-to-beat assessment of cardiac aortic flow. Proc. ISMRM London, 2022.

[26]  Knoll F, Murrell T, Sriram A, Yakubova N, Zbontar J, Rabbat M, et al. Advancing machine learning for MR image reconstruction with an open competition: Overview of the 2019 fastMRI challenge. Magn Reson Med 2020;84:3054–70. https://doi.org/10.1002/mrm.28338.

[27]  Aggarwal HK, Jacob M. J-MoDL: Joint Model-Based Deep Learning for Optimized Sampling and Reconstruction 2019. https://doi.org/10.1109/JSTSP.2020.3004094.

[28]  Bahadir CD, Wang AQ, Dalca A V., Sabuncu MR. Deep-learning-based Optimization of the Under-sampling Pattern in MRI 2019.

[29]  Zhang J, Zhang H, Wang A, Zhang Q, Sabuncu M, Spincemaille P, et al. Extending LOUPE for K-space Under-sampling Pattern Optimization in Multi-coil MRI 2020.

[30]  Jaubert O, Montalt-Tordera J, Knight D, Coghlan GJ, Arridge S, Steeden JA, et al. Real-time deep artifact suppression using recurrent U-Nets for low-latency cardiac MRI. Magn Reson Med 2021;86:1904–16. https://doi.org/10.1002/MRM.28834.